\pdfoutput=1

\documentclass[11pt]{article}

\usepackage[]{acl}

\usepackage{times}
\usepackage{latexsym}

\usepackage[T1]{fontenc}

\usepackage[utf8]{inputenc}

\usepackage{microtype}

\usepackage{inconsolata}

\usepackage{graphicx}

\usepackage{booktabs}
\usepackage{todonotes}
\usepackage{multirow}
\usepackage{comment}
\usepackage{amsmath}
\usepackage{pifont}
\usepackage{inconsolata}
\usepackage{microtype}
\usepackage{amssymb}

\usepackage{amsfonts, amsthm, amssymb}

\title{Datasets for Multilingual Answer Sentence Selection}

\author{Matteo Gabburo$^{1}$\ , Stefano Campese$^{1}$\ ,  Federico Agostini$^{2,3}$\ , Alessandro Moschitti$^{4}$\ \\
$^{1}$University of Trento , $^{2}$Polytechnic University of Turin, $^{3}$University of Padua, $^{4}$Amazon Alexa AI\\
\texttt{\{matteo.gabburo,stefano.campese\}@unitn.it} \\ \texttt{federico.agostini.5@studenti.unipd.it} \\ \texttt{amosch@amazon.com}
}

\begin{document}
\maketitle
\begin{abstract}
Answer Sentence Selection (AS2) is a critical task for designing effective retrieval-based Question Answering (QA) systems. Most advancements in AS2 focus on English due to the scarcity of annotated datasets for other languages. This lack of resources prevents the training of effective AS2 models in different languages, creating a performance gap between QA systems in English and other locales. In this paper, we introduce new high-quality datasets for AS2 in five European languages (French, German, Italian, Portuguese, and Spanish), obtained through supervised Automatic Machine Translation (AMT) of existing English AS2 datasets such as ASNQ, WikiQA, and TREC-QA using a Large Language Model (LLM). We evaluated our approach and the quality of the translated datasets through multiple experiments with different Transformer architectures. The results indicate that our datasets are pivotal in producing robust and powerful multilingual AS2 models, significantly contributing to closing the performance gap between English and other languages.
\end{abstract}

\vspace{-0.3em}
\section{Introduction}
\vspace{-0.3em}
\label{sec:introduction}

Answer Sentence Selection (AS2) represents a crucial component in many QA systems in both academic and industrial settings. The role of this component is to select the correct answer for a given question among a pool of candidate sentences. While in recent years significant progress has been made in developing models and datasets for AS2 \citep{trecqa, wikiqa, tanda,di-liello-etal-2022-paragraph,gupta2023crosslingual}, most of these are designed and evaluated in English. By contrast, less attention has been paid to other medium resource languages, such as French, German, Italian, Portuguese, and Spanish, for which researchers struggle to obtain adequate amounts of quality data to train their models. Recently, Machine Translation (MT) has proven to be an effective approach to address the challenges of low-resource language QA systems \citep{kumar2021machine, ranathunga2023neural, gupta2023crosslingual}.

For the AS2 task, researchers have released a plethora of AS2 datasets in English, such as ASNQ \citep{tanda}, WikiQA \citep{wikiqa}, and TREC-QA \citep{trecqa}, but there remains a gap for lower-resource languages that still needs to be filled.

In this work, we contribute to this research area by introducing three new large multilingual AS2 corpora named mASNQ, mWikiQA, and mTREC-QA for the most common European languages, comprising over 100 million question-answer pairs. We prepared these datasets by translating existing datasets (ASNQ, WikiQA, and TREC-QA) into five European languages (French, German, Italian, Portuguese, and Spanish) using a recent state-of-the-art translation model \citep{nllb}. To validate the effectiveness of our approach, we trained several models using the mASNQ\footnote{\url{https://huggingface.co/datasets/matteogabburo/mASNQ}}, mWikiQA\footnote{\url{https://huggingface.co/datasets/matteogabburo/mWikiQA}}, and mTREC-QA\footnote{\url{https://huggingface.co/datasets/matteogabburo/mTRECQA}} datasets and evaluated their performance. Our results demonstrate that these new datasets can be reliably used to train robust rankers for lower resource languages, yielding higher performance levels than those other competitors achieve. This contribution helps to reduce the language barrier and provides valuable assets for researchers working in low-resource languages.

\vspace{-0.3em}
\section{Related Work}
\vspace{-0.3em}
\label{sec:related_work}

\paragraph{Multilingual Models:} The development of multilingual models has seen significant progress due to the necessity of solving multilingual NLP tasks and cross-lingual applications. mBERT \citep{devlin2019bert}, an extension of the original BERT model, can handle tasks across multiple languages. XLM-RoBERTa \citep{xlm-roberta}, trained on 100 languages, and mDeBERTa \citep{he2021deberta}, a variant of DebertaV3, have shown remarkable improvements in cross-lingual tasks. Similarly, mT5 \citep{mt5}, a multilingual variant of T5, and BLOOM \citep{bloom}, trained on the ROOTS corpus, exemplify advancements in multilingual models. Despite these efforts, multilingual models often underperform compared to their English versions due to the lower availability of training data \citep{gupta2023crosslingual}.

\paragraph{Translation Models:} State-of-the-art Machine Translation (MT) models have demonstrated remarkable capabilities. OPUS-MT \citep{opus-mt}, a set of translation tools, supports both bilingual and multilingual translations. The T5 model \citep{raffel2020exploring}, originally designed for various generative NLP tasks, is widely used for MT. The NLLB model \citep{nllb}, trained on professionally translated datasets, supports translations between over 200 languages, facilitating broader support for low-resource languages.

\paragraph{Machine-Translated Datasets:} MT has been widely used to address the lack of resources for multilingual AS2, showing promising results in the QA domain \citep{vu2021multilingual, kumar2021machine, ranathunga2023neural}. The itSQuAD dataset \citep{squad-ita}, the Spanish SQuAD \citep{carrino2019automatic}, and XQuAD \citep{dumitrescu2021liro} are examples of datasets translated via MT, used to build QA systems in different languages. The MLQA dataset \citep{lewis2019mlqa}, mMARCO \citep{bonifacio2021mmarco}, and Mintaka QA dataset \citep{Sen2022} further highlight the success of machine-translated datasets in QA. Xtr-WikiQA and TyDi-AS2 \citep{gupta2023crosslingual} are recent additions that extend AS2 datasets to multiple languages.

\subsection{Answer Sentence Selection (AS2)}
\label{ssec:as2}

The AS2 task involves selecting the correct sentence from a pool of candidates to answer a given question. Early models like \citet{severyn2016modeling} used separate embeddings for questions and answers, followed by convolutional layers. \citet{tanda} implemented Transformer-based models with an intermediate fine-tuning step, creating the ASNQ corpus from the Natural Questions dataset \citep{nq-data}. Contextual information has been shown to enhance AS2 models \citep{context_as2, Lauriola2021, campese-etal-2023-quadro}. The translation of English AS2 data into target languages has been explored, demonstrating the potential for reducing the complexity of creating multilingual QA systems \citep{DBLP:journals/corr/abs-2102-10250}. Recently, Cross-Lingual Knowledge Distillation (CLKD) \citep{gupta2023crosslingual} has shown impressive results for low-resource languages, although the quality of machine translations remains a critical factor.

\vspace{-0.3em}
\section{AS2 Translated Datasets}
\vspace{-0.3em}
\label{sec:translation}

For dataset translation, we use the largest variant of the NLLB model (NLLB-200-3.3B), which has 3.3 billion parameters. We consider three datasets: TREC-QA, WikiQA, and ASNQ. For each one, we translate both the questions and the answers. Our translation process can be described as a \emph{two-step} procedure. In the first step, we utilize the NLLB model to translate the source datasets into five languages: French, German, Italian, Portuguese, and Spanish. In the second step, we employ two techniques to evaluate the quality of the translations, identifying poor translations, and correcting any misleading sentences.

Firstly, we use a cross-language semantic similarity model released by \citet{reimers-2020-multilingual-sentence-bert}\footnote{\url{https://huggingface.co/sentence-transformers/paraphrase-multilingual-mpnet-base-v2}} to assess the quality of the translated sentences. This model compares the semantic similarity between the original sentences in English and their translated versions in the target language. By measuring the similarity, we can identify bad translations that deviate significantly from the original sentences in English, due to the presence of errors. Secondly, we target these errors by applying a set of heuristics to correct misleading sentences. These heuristics are designed to correct errors, improve clarity, remove non-original text, and enhance the overall quality of the translated datasets.

\vspace{-0.3em}
\subsection{Datasets}
\vspace{-0.3em}

In this work, we considered and translated three datasets for answer sentence selection (AS2) in $5$ different locales:

\textbf{mTREC-QA}, originates from TREC-QA \citep{trecqa}, which is created from the TREC 8 to TREC 13 QA tracks. TREC 8-12 constitutes the training set, while TREC 13 questions are set aside for development and testing. We used the \emph{Clean} setting, meaning that questions without an answer, or with only correct or incorrect answer-sentence candidates are removed.

\textbf{mWikiQA} is the translated version of WikiQA \citep{wikiqa}. It contains 3047 questions sampled from Bing query logs; candidate answer sentences are extracted from Wikipedia, and then manually labeled to assess whether it is a correct answer. Some sentences do not have a correct answer (\emph{all -}), or have only correct answers (\emph{all +}). We trained using \emph{no all -} mode and tested in the \emph{clean} setting (without both \emph{all +} and \emph{all -}).

\textbf{mASNQ} comes from ASNQ \citep{tanda} which is an AS2 dataset created by adapting the Natural Question \citep{nq-data} corpus from Machine Reading (MR) to the AS2 task. We replicated this passage using the scripts provided by \citet{wqa-contextual-qa}.

We summarize the statistics of these datasets in Appendix \ref{apx:datasets_statistics}. 

\subsection{Removing Translation Artifacts}
\label{ssec:bad_translations}

Despite the good quality of the translation, the dataset still presents some inconsistencies and artifacts. We identified four major classes of translation artifacts: (i) Meaning mismatch between the original and the translated sentences, (ii) The addition of not necessary suffixes and prefixes, (iii) The difficulty in interpreting and translating numerical strings, (iv) Out-of-topic translations of partial contexts. We provide some examples of these translation artifacts in Table~\ref{tab:bad_examples}.

\begin{table}
    \small
    \centering
    \resizebox{1.0\linewidth}{!}{%
        \begin{tabular}{@{}ccc@{}}
        
        \toprule
         \textbf{Language}   & \textbf{Split} &    \textbf{Examples}                                       \\ \midrule
        
         \multirow{2}{*}{ITA} & Original & \textit{ISSN 0362 - 4331}                      \\
          & Translated &  ISSN 0362 - 4331 \textcolor{red}{- Non lo so.} \\ \midrule
         \multirow{2}{*}{DEU} & Original  &\textit{Kusumanjali Prakashan .}               \\
         & Translated &\textcolor{red}{Ich bin ein guter Mensch.}     \\ \midrule
         \multirow{2}{*}{FRA} & Original &\textit{Jump up to : `` Safe Haven ( 2013 )}   \\
          &Translated & \textcolor{red}{Sauter sur refuge}             \\ \bottomrule
        \end{tabular}
    }
    \caption{Examples of translation artifacts. The artifacts are highlighted in red. Notice that (i) "Non lo so" is an Italian sentence which translated in English means "I don't know", (ii) "Ich bin ein guter Mensch" means "I am a good person" in German, and (iii) the maening of the "Sauter sur refuge" in French is "Jump on refuge".}
    \label{tab:bad_examples}
    \vspace{-1.5em}
\end{table}

To tackle these issues, we apply some heuristics to improve the dataset quality by designing a simple human-centered pipeline to mitigate these artifacts.

In our approach, we first compute the similarity score between every translated example in the dataset and the corresponding original text. Then we filter out translated examples below a similarity threshold of $0.8$ and, on the remaining set, we compute the most common $1k$ n-grams with $n$ ranging from $4$ to $9$. Second, we manually inspect these extracted n-grams, identifying and removing artifact patterns that could distort the data. Subsequently, we systematically remove occurrences of those problematic artifacts from the translated dataset. To further improve this operation, we also identify the examples where the original sentence and the $75\%$ of the not-blank characters are numbers, and if the similarity score is low (under $0.8$) we replace the translated sentences with the original one.

\vspace{-0.3em}
\subsection{Semantic Similarities}
\vspace{-0.3em}
\label{ssec:semanticsimilarities}

To assess the quality of the translations and to quantify the benefit given by our heuristics, we evaluate the semantic similarity between the original sentences and their translated versions. 
For each question-answer pair of each dataset, we compare the original sentences in English with their translated version in the target language. 
The overall similarity measure between the originals and the translated sentences in each dataset is computed considering the mean of the semantic similarity scores across all the question-answer pairs. This average score indicates how closely the translated sentences align with their original counterparts in terms of semantic meaning. In Appendix~\ref{apx:datasets_statistics} we report the comparison between the similarity scores of the translated dataset and the original one.

\vspace{-0.3em}
\section{Experiments}
\vspace{-0.3em}
\label{sec:experiments}

\begin{table*}[h]
    \small
    \centering
    \resizebox{1\linewidth}{!}{%
    \begin{tabular}{lccccccccc}
        \toprule
        \multirow{2}{*}{\textbf{Language}} & \multirow{2}{*}{\textbf{Transfer}} & \multirow{2}{*}{\textbf{Adapt}} & \multicolumn{2}{c}{\textbf{mWikiQA}} & \multicolumn{2}{c}{\textbf{mTrecQA}} & \multicolumn{2}{c}{\textbf{Xtr-WikiQA}}\\
          &  &   &  \textbf{MAP} & \textbf{P@1} & \textbf{MAP} & \textbf{P@1} & \textbf{MAP} & \textbf{P@1}  \\
         \midrule
         \multirow{2}{*}{ENG}  &            & \checkmark &  0.796 {\tiny($\pm$ 0.011)}            & 0.691 {\tiny($\pm$ 0.015)}             & 0.866 {\tiny($\pm$ 0.015)}             & 0.853 {\tiny($\pm$ 0.031)}             & 0.796 {\tiny($\pm$ 0.011)}             & 0.691 {\tiny($\pm$ 0.015)} \\
                               & \checkmark & \checkmark & \underline{0.874} {\tiny($\pm$ 0.007)} & \underline{0.813} {\tiny($\pm$ 0.017)} & \underline{0.892} {\tiny($\pm$ 0.007)} & \underline{0.871} {\tiny($\pm$ 0.019)} & \underline{0.874} {\tiny($\pm$ 0.007)} & \underline{0.813} {\tiny($\pm$ 0.017)} \\
         \midrule
         \midrule
         \multirow{2}{*}{DEU}  &            & \checkmark & 0.770 {\tiny($\pm$ 0.024)}             & 0.657 {\tiny($\pm$ 0.031)}             & 0.870 {\tiny($\pm$ 0.011)}             & 0.871 {\tiny($\pm$ 0.016)}             & 0.779 {\tiny($\pm$ 0.012)}             & 0.669 {\tiny($\pm$ 0.015)} \\
                               & \checkmark & \checkmark & \underline{0.853} {\tiny($\pm$ 0.005)} & \underline{0.767} {\tiny($\pm$ 0.006)} & \underline{0.904} {\tiny($\pm$ 0.006)} & \underline{0.903} {\tiny($\pm$ 0.017)} & \underline{0.868} {\tiny($\pm$ 0.005)} & \underline{0.800} {\tiny($\pm$ 0.011)} \\  
         \midrule
         \multirow{2}{*}{FRA}  &            & \checkmark & 0.769 {\tiny($\pm$ 0.012)}             & 0.662 {\tiny($\pm$ 0.018)}             & 0.872 {\tiny($\pm$ 0.007)}             & 0.859 {\tiny($\pm$ 0.013)}             & 0.760 {\tiny($\pm$ 0.009)}             & 0.646 {\tiny($\pm$ 0.012)} \\
                               & \checkmark & \checkmark & \underline{0.836} {\tiny($\pm$ 0.006)} & \underline{0.752} {\tiny($\pm$ 0.012)} & \underline{0.891} {\tiny($\pm$ 0.010)} & \underline{0.874} {\tiny($\pm$ 0.029)}             & \underline{0.844} {\tiny($\pm$ 0.005)} & \underline{0.778} {\tiny($\pm$ 0.014)} \\
         \midrule
         \multirow{2}{*}{ITA}  &            & \checkmark & 0.768 {\tiny($\pm$ 0.019)}             & 0.660 {\tiny($\pm$ 0.026)}             & 0.855 {\tiny($\pm$ 0.024)}             & 0.844 {\tiny($\pm$ 0.049)}             & 0.761 {\tiny($\pm$ 0.021)}             & 0.657 {\tiny($\pm$ 0.027)} \\
                               & \checkmark & \checkmark & \underline{0.828} {\tiny($\pm$ 0.004)} & \underline{0.749} {\tiny($\pm$ 0.008)}             & \underline{0.870} {\tiny($\pm$ 0.006)} & \underline{0.871} {\tiny($\pm$ 0.016)} & \underline{0.820} {\tiny($\pm$ 0.012)}             & \underline{0.742} {\tiny($\pm$ 0.027)} \\
         \midrule
         \multirow{2}{*}{POR}  &            & \checkmark & 0.798 {\tiny($\pm$ 0.011)}             & 0.704 {\tiny($\pm$ 0.019)}             & 0.855 {\tiny($\pm$ 0.021)}             & 0.704 {\tiny($\pm$ 0.019)}             & 0.780 {\tiny($\pm$ 0.011)}             & 0.684 {\tiny($\pm$ 0.020)} \\
                               & \checkmark & \checkmark & \underline{0.853} {\tiny($\pm$ 0.018)} & \underline{0.781} {\tiny($\pm$ 0.029)}             & \underline{0.874} {\tiny($\pm$ 0.009)} & \underline{0.781} {\tiny($\pm$ 0.029)}             & \underline{0.849} {\tiny($\pm$ 0.023)} & \underline{0.775} {\tiny($\pm$ 0.042)} \\
         \midrule
         \multirow{2}{*}{SPA}  &            & \checkmark & 0.795 {\tiny($\pm$ 0.009)}             & 0.691 {\tiny($\pm$ 0.016)}             & 0.882 {\tiny($\pm$ 0.013)}             & 0.900 {\tiny($\pm$ 0.016)}             & 0.786 {\tiny($\pm$ 0.015)}             & 0.691 {\tiny($\pm$ 0.024)} \\
                               & \checkmark & \checkmark & \underline{0.847} {\tiny($\pm$ 0.013)} & \underline{0.768} {\tiny($\pm$ 0.020)}             & \underline{0.898} {\tiny($\pm$ 0.004)} & \underline{0.929} {\tiny($\pm$ 0.019)} & \underline{0.859} {\tiny($\pm$ 0.010)} & \underline{0.790} {\tiny($\pm$ 0.020)} \\
         \bottomrule
    \end{tabular}
    }
    \caption{Performance comparison of XLM-RoBERTa on mTREC-QA, mWikiQA, and Xtr-WikiQA (zero-shot from the model trained on mWikiQA). The transfer step is done on mASNQ, while the Adaptation is on mTREC-QA and mWikiQA. Results in terms of MAP and P@1, for various language and model configurations. The experiments on the English split represent the models trained and tested on the original, not translated versions of ASNQ, WikiQA and TREC-QA.}
    \label{tab:results_xmlroberta_mwikiqa_mtrecqa}
    \vspace{-1em}
\end{table*}

In this section, we measure the benefits of our datasets applied to existing multilingual models. With these experiments, we aim to verify and prove the effectiveness of our contributions. Specifically, we want to show that the translated data could be used to train state-of-the-art AS2 models in multiple languages. For each considered language, we finetune existing multilingual transformer models on both the original and our translated datasets. We measure the performance by using information retrieval metrics like Mean Average Precision (MAP) and the Precision at 1 (P@1). These metrics allow us to compare the results with the English baselines trained on the original datasets and to measure the performance improvement given by the ASNQ transfer step on different languages.

To verify these hypotheses, we consider an existing multilanguage pre-trained cross-encoder transformer model, which is XLM-RoBERTa base\footnote{\url{https://huggingface.co/xlm-roberta-base}}, and BERT-multilingual\footnote{\url{https://huggingface.co/bert-base-multilingual-cased}}.
Following the TANDA approach \cite{tanda}, we perform a two-stage training for each model. Precisely, this technique consists of a two-stage training paradigm, where the first training stage, named \emph{transfer step}, involves training the models on ASNQ to teach them to recognize and solve the AS2 tasks. In the second step, named \emph{adaptation step}, the transferred models are fine-tuned on the final target AS2 datasets. In our setting, we apply this paradigm by first training and doing a separate transfer step on each language of mASNQ. Secondly, we finetune the obtained models on mWikiQA and mTREC-QA. 

To have a comprehensive perspective of how our approach behaves, we measure the performance on three test sets from (i) mWikiQA, (ii) mTREC-QA, and (iii) Xtr-WikiQA\footnote{\url{https://huggingface.co/datasets/AmazonScience/xtr-wiki\_qa}}.

With the first two datasets, we aim to show the performance of our models in a controlled environment where the translation pipeline and the heuristics are the same as those used on mASNQ. The third dataset, instead, allows us to prove that (i) our approach is robust in a zero-shot setting, and (ii) can be extended to datasets translated using different pipelines. 

To perform our experiments, we test XLM-RoBERTa on ASNQ and mASNQ datasets with specific parameters: batch size of $1024$, Adam optimizer with a learning rate of $5e-6$, precision set to $32$, and $10$ training epochs. For mWikiQA and mTREC-QA datasets, the batch size was $32$, Adam optimizer with a learning rate of $5e-6$, precision set to 16-mixed, and $40$ training epochs with early stopping. We select the best model maximizing the mean average precision (MAP) on the development set. The same parameters were used for training the Multilingual BERT architecture. All experiments utilized 8 NVIDIA V100 $32$ GB GPUs.

For space reasons, we propose additional experiments using different multilingual models in Appendix~\ref{app:multilingual_better_models_results}.

\vspace{-0.3em}
\subsection{Results}
\label{sssec:as2results}
\vspace{-0.3em}

In this section, we present the experimental results of our approaches on three different AS2 datasets: mWikiQA, mTREC-QA, and the existing Xtr-WikiQA dataset \citep{gupta2023crosslingual}. Table~\ref{tab:results_xmlroberta_mwikiqa_mtrecqa} provides an overview of the performance achieved by our models. First, we observe that our models achieve performance levels comparable to those of English models. This finding is particularly noticeable when considering the Portuguese language across all datasets. When evaluating the models on Xtr-WikiQA, which can be considered as a zero-shot scenario, as the models are trained on mWikiQA and tested on Xtr-WikiQA, we find that our approaches demonstrate robustness even when dealing with datasets translated using a different translation pipeline (Xtr-WikiQA is translated using Amazon Translate). The results obtained on Xtr-WikiQA validate the effectiveness of our procedures in handling such translation variations. 

However, we also observe some negative results. Specifically, our experiments highlight challenges specific to the French language. The XLM-RoBERTa model performance in this context is notably subpar, which aligns with earlier findings documented in the relevant literature. 

\begin{table}[t]
    \small
    \centering
    \resizebox{1.0\linewidth}{!}{%
    \begin{tabular}{llcccc}
        \toprule
        \textbf{Language} & \textbf{Model} & \textbf{MAP} & \textbf{P@1} \\
        \midrule
        \multirow{4}{*}{ENG} & \ding{81} mmarco-mMiniLMv2                       & 0.812 &  0.722 \\
        & \ding{81} bert-multilingual-msmarco                                   & 0.798 &  0.714 \\
        & xlm-roberta-base                                                           & \underline{0.855} &  \underline{0.794} \\
        & bert-multilingual                                                     & 0.814 &  0.724 \\
        \midrule
        \multirow{4}{*}{DEU}     &\ding{81} mmarco-mMiniLMv2   & 0.797 &  0.700 \\
        &\ding{81} bert-multilingual-msmarco                   & 0.759 &  0.663 \\
        &xlm-roberta-base                                           & \underline{0.844} &  \underline{0.770} \\
        &bert-multilingual                                     & 0.834 &  0.753 \\
        \midrule
        \multirow{4}{*}{FRA}     &\ding{81} mmarco-mMiniLMv2   & 0.782 &  0.675 \\
        &\ding{81} bert-multilingual-msmarco                   & 0.734 &  0.621 \\
        & xlm-roberta-base                                          & 0.813 &  0.720 \\
        & bert-multilingual                                    & \underline{0.863} &  \underline{0.807} \\
        \midrule
        \multirow{4}{*}{ITA}     & \ding{81} mmarco-mMiniLMv2  & 0.778 &  0.671 \\
        & \ding{81} bert-multilingual-msmarco                  & 0.735 &  0.634 \\
        & xlm-roberta-base                                          & 0.830 &  0.741 \\
        & bert-multilingual                                    & \underline{0.846} &  \underline{0.765} \\
        \midrule
        \multirow{4}{*}{POR}     &\ding{81} mmarco-mMiniLMv2    & 0.809 & 0.724 \\
        & \ding{81} bert-multilingual-msmarco                   & 0.755 & 0.646 \\
        & xlm-roberta-base                                           & 0.840 & \underline{0.761} \\
        & bert-multilingual                                     & \underline{0.841} & \underline{0.761} \\
        \midrule
        \multirow{4}{*}{SPA}     &\ding{81} mmarco-mMiniLMv2    & 0.791 & 0.691 \\
        & \ding{81} bert-multilingual-msmarco                   & 0.753 & 0.650 \\
        & xlm-roberta-base                                           & 0.832 & 0.737 \\
        & bert-multilingual                                     & \underline{0.853} & \underline{0.774} \\
         \bottomrule
    \end{tabular}
    }
    \caption{Performance comparison of XLM-RoBERTa base model in a zero-shot setting on the Xtr-WikiQA task. Models trained on mASNQ dataset, denoted by \ding{81}, outperform those trained on other datasets like mMARCO and MSMARCO. Moreover, BERT-multilingual consistently performs better than XLM-RoBERTa in various languages (Italian, Portuguese, Spanish), indicating the robustness and competitiveness of the approach on AS2 datasets.}
    \label{tab:results_xml_wikiqa_xtr_zeroshot}
    \vspace{-1.5em}
\end{table}

In addition, we present a comparison of various models on Xtr-WikiQA in a zero-shot setting in Table~\ref{tab:results_xml_wikiqa_xtr_zeroshot}. These models have been trained on mASNQ and are evaluated against existing models trained on well-known and extensive passage reranking datasets. We find that models trained on mASNQ for the Xtr-WikiQA task outperform models trained on other datasets such as mMARCO and MSMARCO. This observation suggests that mASNQ is a more suitable dataset for AS2 compared to mMARCO and MSMARCO. Moreover, when comparing the performance of BERT-multilingual and XLM-RoBERTa, we find that, on average, BERT-multilingual performs better. This finding is evident when analyzing the results across different languages, including Italian, Portuguese, and Spanish. Overall, our results demonstrate the effectiveness and robustness of our approaches on AS2 datasets, showcasing competitive performance and the superiority of certain training configurations and models over others.

\vspace{-0.3em}
\section{Ablation Studies}
\label{sec:appendix_ablations}
\vspace{-0.3em}


We conducted several experiments to estimate the benefits provided by our multilingual datasets, assessing their impact on different aspects of model performance.
\paragraph{Cross-Lingual:} Models trained on the mASNQ dataset consistently outperformed those trained on ASNQ, demonstrating higher MAP and P@1 scores across all languages. This confirms the effectiveness of mASNQ in enhancing cross-lingual model performance (Appendix~\ref{apx:cross_lingual_apx}).
\paragraph{Ranks Correlation:} Models trained on mASNQ and mWikiQA showed strong positive correlations in their ranking outputs compared to those trained on ASNQ and WikiQA. This indicates consistent translation quality and robust model performance (Appendix~\ref{apx:ranks_correlation}).
\paragraph{Passage Ranking:} Models trained on mMARCO outperformed those trained on MSMARCO, emphasizing the significant advantages provided by adapting models trained on our multilingual datasets for various tasks (Appendix~\ref{apx:passagereranking}).

\vspace{-0.3em}
\section{Conclusion}
\vspace{-0.3em}
\label{sec:conclusion}

Our study tackles the language barrier in QA systems by focusing on European languages such as Italian, German, Portuguese, Spanish, and French. We introduced new large multilingual AS2 datasets (mASNQ, mWikiQA, and mTREC-QA) by translating existing English AS2 datasets using a state-of-the-art translation model. This approach provides valuable resources for lower-resource languages.
Our extensive experiments demonstrated the effectiveness of these datasets in training robust AS2 rankers across various languages, achieving performance comparable to English datasets. This contributes significantly to reducing the language barrier, making AS2 more accessible and effective across different linguistic contexts.
To support further research, we will release the new models and multilingual AS2 datasets to the research community. We hope our work inspires future studies to address language diversity challenges in QA, leading to more inclusive and effective solutions for global users.




\clearpage

\section*{Limitations}
This paper focuses on five European languages (Italian, German, Portuguese, Spanish, and French). This could represent a limitation since we limit the applicability of the findings to other languages. Another possible limitation is that the accuracy and quality of machine translation can affect the performance of trained models by introducing errors and inconsistencies, compromising dataset reliability. Moreover, biases present in the original English data might be transferred to the translated datasets, potentially resulting in skewed or unrepresentative training examples for specific languages. Finally, we reserve for future analysis on larger and more powerful pre-trained multilingual language models (e.g., XLM-RoBERTa large, and mDeBERTa).

\nocite{*}

\bibliography{custom}


\appendix

\section{ASNQ additional results}
\label{apx:asnq_additional_results}

Table~\ref{tab:asnq_xlm_roberta_results} presents the performance of the XLM-RoBERTa model trained on the development set of the multilingual ASNQ dataset. The performance of XLM-RoBERTa on the original ASNQ development set is also reported. The results indicate that the English baseline outperforms the models trained in other languages, as expected, while the performance of the models trained in different languages is consistent and relatively close. These results highlight that the use of our translated datasets can improve the performance in terms of MAP, P@1, MRR, and NDCG metrics across multiple languages.

\begin{table*}[htp]
    \small
    \centering
    \resizebox{1.0\linewidth}{!}{%
        \begin{tabular}{lcccccccc}
        \toprule
        \multirow{2}{*}{\textbf{Language}} & \multicolumn{4}{c}{\textbf{With Translated Data}} & \multicolumn{4}{c}{\textbf{Without Translated Data}}\\
          &  \textbf{MAP} & \textbf{P@1} & \textbf{MRR} & \textbf{NDCG} & \textbf{MAP} & \textbf{P@1} & \textbf{MRR} & \textbf{NDCG} \\
         \midrule
         ENG  & 0.870 {\tiny($\pm$ 0.007)} & 0.812 {\tiny($\pm$ 0.017)} & 0.745 {\tiny($\pm$ 0.011)} & 0.831 {\tiny($\pm$ 0.015)} & 0.870 {\tiny($\pm$ 0.007)} & 0.812 {\tiny($\pm$ 0.017)} & 0.745 {\tiny($\pm$ 0.011)} & 0.831 {\tiny($\pm$ 0.015)} \\
         \midrule
         DEU   & 0.850 {\tiny($\pm$ 0.005)} & 0.768 {\tiny($\pm$ 0.006)} & 0.710 {\tiny($\pm$ 0.008)} & 0.811 {\tiny($\pm$ 0.009)} & 0.845 {\tiny($\pm$ 0.005)} & 0.762 {\tiny($\pm$ 0.010)} & 0.705 {\tiny($\pm$ 0.010)} & 0.806 {\tiny($\pm$ 0.012)} \\
         FRA   & 0.835 {\tiny($\pm$ 0.006)} & 0.750 {\tiny($\pm$ 0.012)} & 0.692 {\tiny($\pm$ 0.009)} & 0.798 {\tiny($\pm$ 0.011)} & 0.830 {\tiny($\pm$ 0.010)} & 0.743 {\tiny($\pm$ 0.017)} & 0.689 {\tiny($\pm$ 0.014)} & 0.793 {\tiny($\pm$ 0.017)} \\
         ITA  & 0.842 {\tiny($\pm$ 0.004)} & 0.755 {\tiny($\pm$ 0.008)} & 0.699 {\tiny($\pm$ 0.009)} & 0.804 {\tiny($\pm$ 0.010)} & 0.835 {\tiny($\pm$ 0.003)} & 0.748 {\tiny($\pm$ 0.011)} & 0.692 {\tiny($\pm$ 0.011)} & 0.798 {\tiny($\pm$ 0.012)} \\
         POR & 0.853 {\tiny($\pm$ 0.018)} & 0.781 {\tiny($\pm$ 0.029)} & 0.715 {\tiny($\pm$ 0.021)} & 0.822 {\tiny($\pm$ 0.023)} & 0.848 {\tiny($\pm$ 0.011)} & 0.777 {\tiny($\pm$ 0.020)} & 0.712 {\tiny($\pm$ 0.015)} & 0.818 {\tiny($\pm$ 0.017)} \\
         SPA  & 0.847 {\tiny($\pm$ 0.013)} & 0.768 {\tiny($\pm$ 0.020)} & 0.705 {\tiny($\pm$ 0.016)} & 0.812 {\tiny($\pm$ 0.018)} & 0.840 {\tiny($\pm$ 0.012)} & 0.760 {\tiny($\pm$ 0.024)} & 0.700 {\tiny($\pm$ 0.018)} & 0.805 {\tiny($\pm$ 0.019)} \\
         \bottomrule
        \end{tabular}
    }
    \caption{Performance comparison of XLM-RoBERTa on the multilingual ASNQ dataset with and without translated data. Results are reported in terms of MAP, P@1, MRR, and NDCG metrics.}
    \label{tab:asnq_xlm_roberta_results}
\end{table*} 

\section{Results using better multilingual models}
\label{app:multilingual_better_models_results}

In this section, we present the results of our experiments using the newly created multilingual Answer Sentence Selection (AS2) datasets. The goal is to evaluate the performance of our approach using mDeBERTa across different languages and settings. We consider three main tables that provide a comprehensive overview of the results.

Table~\ref{tab:results_masnq} presents the performance of mDeBERTa on the mASNQ dataset, covering multiple languages (DEU, FRA, ITA, SPA, POR). The metrics reported include Mean Average Precision (MAP), Mean Reciprocal Rank (MRR), Normalized Discounted Cumulative Gain (NDCG), and Precision at 1 (P@1). These results highlight the effectiveness of our multilingual datasets, showcasing the robustness and consistency of the model across different languages.

\begin{table}[htp]
    \small
    \centering
    \begin{tabular}{lcccc}
        \toprule
         Approach & MAP & MRR & NDCG & P1 \\
         \midrule
         ASNQ                           & 0.677 & 0.743 &	0.707	& 0.638 \\
         \midrule
         mASNQ$_{DEU}$                    & 0.633 & 0.702 & 0.662	& 0.590 \\
         mASNQ$_{FRA}$                    & 0.629 & 0.700 & 0.657	& 0.591 \\
         mASNQ$_{ITA}$                    & 0.639 & 0.708 & 0.670   & 0.597 \\
         mASNQ$_{SPA}$                    & 0.632 & 0.703 & 0.663	& 0.589 \\
         mASNQ$_{POR}$                    &  0.635 & 0.706 & 0.664	& 0.595 \\
         \bottomrule
    \end{tabular}
    \caption{Results of mDeberta on mASNQ}
    \label{tab:results_masnq}
\end{table}


Table~\ref{tab:results_deberta_crosslingual} compares the performance of mDeBERTa-v3-base when transferred on mASNQ and ASNQ, and subsequently tested on mWikiQA. The results are presented in terms of MAP and P@1 for each language considered (ITA, DEU, SPA, POR, FRA). This table demonstrates the improvements achieved by utilizing the mASNQ dataset, with notable gains in performance across all languages.

\begin{table}[hpt]
\centering
\small
\begin{tabular}{@{}lcccc@{}}
\toprule
    & \multicolumn{2}{c}{ASNQ} & \multicolumn{2}{c}{mASNQ} \\
    & MAP         & P@1        & MAP         & P@1         \\
\midrule
ITA &  0.868 & 0.801 &  0.884 & 0.821 \\
DEU &  0.882 & 0.827 &  0.885 & 0.821 \\
SPA &  0.875 & 0.811 &  0.886 & 0.829 \\
POR &  0.882 & 0.825 &  0.883 & 0.830 \\
FRA &  0.851 & 0.766 &  0.884 & 0.819 \\
\bottomrule
\end{tabular}
\caption{Results of mDeBERTa-v3-base transferred on mASNQ and ASNQ and tested on mWikiQA.}
\label{tab:results_deberta_crosslingual}
\end{table}


Table~\ref{tab:combined_results_deberta} presents a detailed performance comparison of mDeBERTa on three datasets: mWikiQA, mTREC-QA, and Xtr-WikiQA (zero-shot from the model trained on mWikiQA). The results are reported in terms of MAP and P@1 for various language and model configurations. This table illustrates the benefits of the transfer step on mASNQ and the adaptation step on mTREC-QA and mWikiQA, with the mDeBERTa models consistently achieving high performance across all tasks and languages.

\begin{table*}[htp]
    \small
    \centering
    \resizebox{1.0\linewidth}{!}{%
        \begin{tabular}{lcccccccccc}
        \toprule
        \multirow{2}{*}{\textbf{Language}} & \multirow{2}{*}{\textbf{Transfer}} & \multirow{2}{*}{\textbf{Adapt}} & \multicolumn{2}{c}{\textbf{mWikiQA}} & \multicolumn{2}{c}{\textbf{mTREC-QA}} & \multicolumn{2}{c}{\textbf{Xtr-WikiQA}} \\
          &  &   &  \textbf{MAP} & \textbf{P@1} & \textbf{MAP} & \textbf{P@1} & \textbf{MAP} & \textbf{P@1}  \\
         \midrule

         \multirow{2}{*}{ENG} &            & \checkmark &                                    0.872   {\tiny($\pm$  0.008)} &             0.801   {\tiny($\pm$  0.013)} &             0.905   {\tiny($\pm$  0.004)} &             0.909   {\tiny($\pm$  0.022)}  &            0.872   {\tiny($\pm$  0.008)} &             0.801   {\tiny($\pm$  0.013)} \\
                               & \checkmark & \checkmark  & \underline{0.901} {\tiny($\pm$  0.006)} & \underline{0.854} {\tiny($\pm$  0.008)} & \underline{0.921} {\tiny($\pm$  0.003)} & \underline{0.950} {\tiny($\pm$  0.008)} & \underline{0.901} {\tiny($\pm$  0.006)} & \underline{0.854} {\tiny($\pm$  0.008)} \\  
         \midrule
         \midrule
         \multirow{2}{*}{DEU}  &            & \checkmark & 0.847 {\tiny($\pm$ 0.006)} & 0.765 {\tiny($\pm$ 0.006)} & 0.898 {\tiny($\pm$ 0.011)} & 0.912 {\tiny($\pm$ 0.023)} & 0.847 {\tiny($\pm$ 0.006)} & 0.765 {\tiny($\pm$ 0.006)} \\
                               & \checkmark & \checkmark & \underline{0.888} {\tiny($\pm$ 0.007)} & \underline{0.837} {\tiny($\pm$ 0.013)} & \underline{0.925} {\tiny($\pm$ 0.008)} & \underline{0.944} {\tiny($\pm$ 0.016)} & \underline{0.888} {\tiny($\pm$ 0.007)} & \underline{0.837} {\tiny($\pm$ 0.013)} \\  
         \midrule
         \multirow{2}{*}{FRA}  &            & \checkmark & 0.848 {\tiny($\pm$ 0.010)} & 0.778 {\tiny($\pm$ 0.018)} & 0.900 {\tiny($\pm$ 0.008)} & 0.924 {\tiny($\pm$ 0.016)} &                                                            0.848  {\tiny($\pm$ 0.010)} &            0.778  {\tiny($\pm$ 0.018)} \\
                               & \checkmark & \checkmark & \underline{0.894} {\tiny($\pm$ 0.003)} & \underline{0.848} {\tiny($\pm$ 0.003)} & \underline{0.918} {\tiny($\pm$ 0.004)} & \underline{0.932} {\tiny($\pm$ 0.008)} & \underline{0.894} {\tiny($\pm$ 0.003)} & \underline{0.848} {\tiny($\pm$ 0.003)} \\
         \midrule
         \multirow{2}{*}{ITA}  &            & \checkmark & 0.851 {\tiny($\pm$ 0.010)} & 0.774 {\tiny($\pm$ 0.011)} & 0.890 {\tiny($\pm$ 0.009)} & 0.900 {\tiny($\pm$ 0.016)} &                                                            0.851  {\tiny($\pm$ 0.010)} &            0.774  {\tiny($\pm$ 0.011)} \\
                               & \checkmark & \checkmark & \underline{0.885} {\tiny($\pm$ 0.006)} & \underline{0.822} {\tiny($\pm$ 0.010)} & \underline{0.919} {\tiny($\pm$ 0.005)} & \underline{0.938} {\tiny($\pm$ 0.012)} & \underline{0.885} {\tiny($\pm$ 0.006)} & \underline{0.822} {\tiny($\pm$ 0.010)} \\
         \midrule
         \multirow{2}{*}{POR}  &            & \checkmark & 0.846 {\tiny($\pm$ 0.015)} & 0.765 {\tiny($\pm$ 0.023)} & 0.896 {\tiny($\pm$ 0.013)} & 0.900 {\tiny($\pm$ 0.019)} &                                                            0.846  {\tiny($\pm$ 0.015)} &            0.765  {\tiny($\pm$ 0.023)} \\
                               & \checkmark & \checkmark & \underline{0.889} {\tiny($\pm$ 0.005)} & \underline{0.842} {\tiny($\pm$ 0.007)} & \underline{0.925} {\tiny($\pm$ 0.004)} & \underline{0.953} {\tiny($\pm$ 0.012)} & \underline{0.889} {\tiny($\pm$ 0.005)} & \underline{0.842} {\tiny($\pm$ 0.007} \\
         \midrule
         \multirow{2}{*}{SPA}  &            & \checkmark & 0.857 {\tiny($\pm$ 0.010)} & 0.781 {\tiny($\pm$ 0.016)} & 0.902 {\tiny($\pm$ 0.012)} & 0.921 {\tiny($\pm$ 0.022)} &                                                            0.857  {\tiny($\pm$ 0.010)} &            0.781  {\tiny($\pm$ 0.016)} \\
                               & \checkmark & \checkmark & \underline{0.879} {\tiny($\pm$ 0.007)} & \underline{0.819} {\tiny($\pm$ 0.011)} & \underline{0.915} {\tiny($\pm$ 0.007)} & \underline{0.924} {\tiny($\pm$ 0.019)} & \underline{0.879} {\tiny($\pm$ 0.007)} & \underline{0.819} {\tiny($\pm$ 0.011)} \\
         \bottomrule
    \end{tabular}
   }
    \caption{Performance comparison of mDeBERTa on mWikiQA, mTREC-QA, and Xtr-WikiQA (zero-shot from the model trained on mWikiQA). The transfer step is done on mASNQ, while the adaptation is on mTREC-QA and mWikiQA. Results in terms of MAP and P@1, for various language and model configurations.}
    \label{tab:combined_results_deberta}
\end{table*}

The results in these tables provide comprehensive insights into the effectiveness of our multilingual datasets and the benefits of the proposed transfer and adaptation steps. These findings underline the importance of high-quality multilingual datasets in improving the performance of AS2 models across diverse languages, demonstrating the robustness and generalizability of our approach.

\section{Ablation: Cross-Lingual}
\label{apx:cross_lingual_apx}

This ablation aims to determine the advantages of using the mASNQ dataset to train state-of-the-art answer ranking models on languages different from English. To achieve this, we compare the performance of cross-lingual models trained on ASNQ and WikiQA with models that were first trained on mASNQ and mWikiQA, across the different languages that compose mWikiQA.  

Table~\ref{tab:results_xml_crosslingual} compares the performance of models trained only on the original versions of ASNQ and WikiQA with the performance of the same architecture (XLM-RoBERTa base) but trained on our multilingual datasets. To achieve this goal, we measure the performance of each model across all the different test sets of mWikiQA and across their languages. For the evaluation, we considered two proxy measures to understand the quality of the models: Mean Average Precision (MAP) and Precision at 1 (P@1). The results show that the models achieve higher MAP and P@1 scores when trained on mASNQ compared to ASNQ, indicating that training on the mASNQ dataset improves the performance of multilingual models in cross-lingual tasks. Across all languages, the models trained on mASNQ consistently outperform the models trained on ASNQ. This suggests that the mASNQ dataset can guarantee a performance boost for non-English target datasets, confirming our hypotheses.

\begin{table}[hpt]
\centering
\small
\resizebox{0.85\linewidth}{!}{%
\begin{tabular}{@{}lcccc@{}}
\toprule
\multirow{2}{*}{\textbf{Language}}   & \multicolumn{2}{c}{\textbf{ASNQ}} & \multicolumn{2}{c}{\textbf{mASNQ}} \\
    & \textbf{MAP}         & \textbf{P@1}        & \textbf{MAP}         & \textbf{P@1}         \\
\midrule
DEU &  0.814 & 0.705 &  0.839 & 0.755 \\
FRA &  0.819 & 0.717 &  0.793 & 0.671 \\
ITA &  0.819 & 0.715 &  0.839 & 0.726 \\
POR &  0.822 & 0.722 &  0.842 & 0.755 \\
SPA &  0.830 & 0.738 &  0.835 & 0.751 \\
\bottomrule
\end{tabular}}
\caption{Comparison of XLM-RoBERTa base transferred on mASNQ and ASNQ and tested on mWikiQA in a cross-lingual setting.}
\label{tab:results_xml_crosslingual}
\vspace{-0.6em}
\end{table}

\section{Ablation: Ranks Correlation}
\label{apx:ranks_correlation}

This study compares the ranking outputs of two sets of models, analyzing the correlation between their rankings. The first set comprehends models trained on mASNQ and mWikiQA and then tested on the mWikiQA test set, while the second set contains models trained on ASNQ and WikiQA and evaluated on the original English WikiQA test set.

We design this experiment in order to compare the rank provided for each question $q_{Eng}^{i}$ of the original English dataset (WikiQA), with the semantically equivalent question $q_{T}^{i}$ and its rank for each language $T$ in mWikiQA. To measure the performance, we compute three correlation metrics to properly evaluate the correlation between the rankings of each pair of questions $\{q_{Eng}^{i}, q_{T}^{i}\}$; in this way, we allow determining the level of agreement between the two models' ranking outputs, providing insights into the potential differences between them. Specifically, we consider XLM-RoBERTa base and compute the Kendall, Spearman, and Pearson correlation metrics on mWikiQA and mTREC-QA. 

The results in Table~\ref{tab:resultRanksCorrelation} show a strong positive correlation between the performance of models trained in English and tested in English, and the models trained in other languages (using mASNQ, mWikiQA, and mTREC-QA). This correlation is evident across all evaluation metrics, with Kendall correlations ranging from $0.694$ to $0.720$, Spearman correlations ranging from $0.802$ to $0.824$, and Pearson correlations ranging from $0.872$ to $0.908$ for the mASNQ$\rightarrow$mWikiQA task. The high correlation values, ranging from $0.547$ to $0.733$, across all languages for the mASNQ$\rightarrow$mTREC-QA task further support this notion. The Kendall, Spearman, and Pearson correlations show consistently high values, indicating that the translation quality and model performance are consistently strong. The results of the analysis demonstrate (i) the effectiveness of the translation process for mASNQ and (ii) the strong performance of the models.

\begin{table}[hpt]
\small
\centering
\resizebox{1.0\linewidth}{!}{
\begin{tabular}{cccc}
\toprule
\multicolumn{4}{c}{\textbf{mASNQ$\rightarrow$mWikiQA}} \\ 
\midrule
\textbf{Language} & \textbf{Kendall} & \textbf{Spearman} & \textbf{Pearson} \\ 
\midrule
DEU & 0.720 & 0.820 & 0.908 \\
FRA & 0.694 & 0.802 & 0.872 \\
ITA & 0.713 & 0.817 & 0.903 \\
POR & 0.710 & 0.824 & 0.908 \\
SPA & 0.698 & 0.807 & 0.902 \\
\midrule
\end{tabular}}
\resizebox{1.0\linewidth}{!}{%
\begin{tabular}{cccc}
\multicolumn{4}{c}{\textbf{mASNQ$\rightarrow$mTREC-QA}} \\ 
\midrule
\textbf{Language} & \textbf{Kendall} & \textbf{Spearman} & \textbf{Pearson} \\ 
\midrule
DEU & 0.547 & 0.666 & 0.713 \\
FRA & 0.566 & 0.663 & 0.709 \\
ITA & 0.513 & 0.629 & 0.695 \\
POR & 0.567 & 0.669 & 0.733 \\
SPA & 0.587 & 0.688 & 0.728 \\ 
\bottomrule
\end{tabular}}
\caption{Kendall, Spearman and Pearson correlation computed between the ranks originated from model trained the original ASNQ and mASNQ. The reported values are computed using XLM-RoBERTa base models transferred on ASNQ and mASNQ and then finetuned on mWikiQA and mTREC-QA.}
\label{tab:resultRanksCorrelation}
\end{table}

\section{Ablation: Passage Ranking}
\label{apx:passagereranking}

To further evaluate the robustness of our datasets, we also perform several experiments on a different task: Passage Reranking (PR). Passage Reranking is an Information Retrieval (IR) task that consists of reordering a set of retrieved passages for a given query. For this reason, we consider a well-known dataset named mMARCO \citep{bonifacio2021mmarco}, well known in the multilingual IR community. Specifically, we select a random language among the ones considered in the previous experiments, and we train several multi-language models. In detail, we split the original Italian dataset into train, validation, and test splits (Tab.~\ref{tab:datasets_statistics}).

We compare the results obtained by our approaches with two models: the first is a multilingual BERT trained on the English MSMARCO, while the second model is trained on our train split. In Table~\ref{tab:results_mmarco}, we present the results of this comparison. They clearly show that our models trained on the mMARCO dataset outperform the model trained on MSMARCO (e.g., $0.687$ vs $0.682$ in terms of MAP). 

Although the improvement is modest, it becomes significant due to the large size of the mMARCO test set. These findings highlight the advantages our datasets offer for tasks beyond AS2. Even with a marginal improvement, it is evident that adapting a model trained on our multilingual datasets can yield further performance enhancements.

\begin{table}[htp]
\vspace{0.5em}
    \small
    \centering
    \resizebox{1.0\linewidth}{!}{%
        \begin{tabular}{lcccc}
        \toprule
        \multirow{2}{*}{\textbf{Dataset}} & \multirow{2}{*}{\textbf{Transfer}} & \multirow{2}{*}{\textbf{Adapt}} & \multicolumn{2}{c}{\textbf{mMARCO}}\\
          &  &   &  \textbf{MAP} & \textbf{P@1}  \\
             \midrule 
             MSMARCO                            & \checkmark  &   &  0.631	           & 0.502 \\
             mMARCO                     &    & \checkmark & 0.682	           & 0.553 \\
             \midrule
             mASNQ$\xrightarrow{}$mMARCO     & \checkmark & \checkmark & \underline{0.687} & \underline{0.559} \\
             \bottomrule
        \end{tabular}
    }    
    \caption{Comparison of BERT-multilingual performance on mMARCO$_\text{ITA}$ test set. We train the two baselines respectively on the English MSMARCO and the mMARCO Italian split. The models trained on mASNQ and adapted to mMARCO consistently improve the two presented baselines, showing that the transfer step on mASNQ is helpful in this domain.} 
    \label{tab:results_mmarco}
\end{table}

\section{Datasets}
\label{apx:datasets_statistics}

In Table~\ref{tab:datasets_statistics}, we provide the datasets we described in Section \ref{sec:translation}. 
\begin{table}[htp]
    \small
    \centering
    \resizebox{\linewidth}{!}{
    \begin{tabular}{llccc}
        \toprule
         \textbf{Dataset} & \textbf{Split} & \textbf{\#Question} & \textbf{\#QA Pairs} \\
         \midrule
         \multirow{2}{*}{mASNQ}          & Train      & 57240 & 20377168 \\
                                        & Validation & 2672 & 930062 \\
         \midrule
        \multirow{7}{*}{mWikiQA}                         & Train      & 2118 & 20356\\
                                        & Validation & 296 & 2731 \\
                                        & Validation (++) & 126 & 1130 \\
                                        & Validation (clean) & 122 & 1126 \\
                                        & Test       & 633 & 6160 \\
                                        & Test (++)   & 243 & 2350 \\
                                        & Test (clean) & 237 & 2340 \\
        \midrule
         \multirow{3}{*}{mTREC-QA}      & Train      & 1227 & 53282 \\
                                        & Validation & 65 & 1117 \\
                                        & Test       & 68 & 1441 \\
         \bottomrule
    \end{tabular}}
    \caption{Dataset statistics for mASNQ, mWikiQA, and mTREC-QA for each language. The datasets have the same statistics in their original version, and considering all the languages, the corpora comprehend more than 100M examples. Notice that for mWikiQA we report also the statistics of the clean and the no-all-negatives (++) splits.}
    \label{tab:datasets_statistics}
    \vspace{-1em}
\end{table}

In addition, in Table \ref{tab:similarities_asnq_masnq}, we report the semantic similarity between ASNQ and mASNQ to support the translation quality further. 
\begin{table}[htp]
    \small
    \centering
    \resizebox{1.0\linewidth}{!}{%
    \begin{tabular}{ccc}
        \toprule
        & \textbf{Language} & \textbf{Similarity} \\
        \midrule
        \multirow{5}{*}{\rotatebox{90}{Dev}}& DEU & $0.869\pm0.178\rightarrow0.991\pm0.003$ \\
                                            & FRA & $0.838\pm0.223\rightarrow0.990\pm0.003$ \\
                                            & ITA & $0.913\pm0.115\rightarrow0.990\pm0.001$ \\
                                            & POR & $0.915\pm0.117\rightarrow0.992\pm0.003$ \\
                                            & SPA & $0.857\pm0.211\rightarrow0.990\pm0.001$ \\

        \midrule
        \multirow{5}{*}{\rotatebox{90}{Train}} & DEU & $0.871\pm0.175\rightarrow0.923\pm0.110$ \\
                                               & FRA & $0.841\pm0.218\rightarrow0.923\pm0.101$ \\
                                               & ITA & $0.915\pm0.112\rightarrow0.940\pm0.081$ \\
                                               & POR & $0.915\pm0.115\rightarrow0.941\pm0.082$ \\
                                               & SPA & $0.860\pm0.206\rightarrow0.932\pm0.090$ \\
        \bottomrule
    \end{tabular}}
    \caption{Similarities between ASNQ and mASNQ. On the left of the arrow ($\rightarrow$) the similarity reached after the initial translation is reported; on the right side, there is the similarity score after the application of the heuristics.}
    \label{tab:similarities_asnq_masnq}
\end{table}


\end{document}